\documentclass[11pt,a4paper]{article}
\usepackage[hyperref]{acl2019}
\usepackage{times}
\usepackage{latexsym}
\usepackage{graphicx}
\usepackage{subcaption}
\usepackage{amsmath}
\usepackage{xcolor}
\usepackage{enumitem}
\newcommand{\bx}{\mathbf{x}}
\newcommand{\bh}{\mathbf{h}}

\usepackage{url}

\usepackage{cite}
\usepackage{amsmath,amssymb,amsfonts}
\usepackage{algorithmic}
\usepackage{graphicx}
\usepackage{textcomp}
\usepackage{xcolor}
\def\BibTeX{{\rm B\kern-.05em{\sc i\kern-.025em b}\kern-.08em
    T\kern-.1667em\lower.7ex\hbox{E}\kern-.125emX}}

\usepackage{times}
\usepackage{latexsym}
\usepackage{graphicx}
\usepackage{amsfonts}
\usepackage{amsmath,amsthm,amssymb,scrextend}
\usepackage{bm}
\usepackage{bbm}
\usepackage{booktabs}
\usepackage{array}
\usepackage{verbatim}
\usepackage[]{natbib}

\aclfinalcopy 

\title{Assessing incrementality in sequence-to-sequence models}

\author{
  Dennis Ulmer \\
  University of Amsterdam \\
  \texttt{dennis.ulmer@gmx.de}
    \\\And
  Dieuwke Hupkes \\
  ILLC, University of Amsterdam \\
  \texttt{d.hupkes@uva.nl} \\\And
  Elia Bruni \\
  Universitat Pompeu Fabra \\
  \texttt{elia.bruni@gmail.com} \\
  }

\date{}

\begin{document}
\maketitle
\begin{abstract}
  Since their inception, encoder-decoder models have successfully been applied to a wide array of problems in computational linguistics. The most recent successes are predominantly due to the use of different variations of attention mechanisms, but their cognitive plausibility is questionable.
  In particular, because past representations can be revisited at any point in time, attention-centric methods seem to lack an incentive to build up incrementally more informative representations of incoming sentences.
  This way of processing stands in stark contrast with the way in which humans are believed to process language: continuously and rapidly integrating new information as it is encountered.
  In this work, we propose three novel metrics to assess the behavior of RNNs with and without an attention mechanism and identify key differences in the way the different model types process sentences.
\end{abstract}

\section{Introduction}


 Incrementality -- that is, building up representations ``as rapidly as possible as the input is encountered'' \citep{christiansen2016now} -- is considered one of the key ingredients for humans to process language efficiently and effectively.

\citet{christiansen2016now} conjecture how this trait is realized in human cognition by identifying several components which either make up or are implications of their hypothesized \emph{Now-or-Never bottleneck}, a set
of fundamental constraints on human language processing, which include a limited amount of available memory and time pressure.
First of all, one of the implications of the now-or-never bottleneck is anticipation, implemented by a mechanism called \emph{predictive processing}.
As humans have to process sequences of inputs fast, they already try to anticipate the next element before it is being uttered. This is hypothesized to be the reason why people struggle with so-called garden path sentences like ``The horse race past the barn fell", where the last word encountered, ``fell'', goes against the representation of the sentence built up until this point.
Secondly, another strategy being employed by humans in processing language seems to be \emph{eager processing}: the cognitive system encodes new input into ``rich'' representations as fast as possible. These are build up in chunks and then processed into more and more abstract representations, an operation
\citet{christiansen2016now} call \emph{Chunk-and-pass processing}.


In this paper, we aim to gain a better insight into the inner workings of recurrent models with respect to incrementality while taking inspiration from and drawing parallels to this psycholinguistic perspective. To ensure a successful processing of language, the human brain seems to be forced to employ an encoding scheme that seems highly reminiscent of the encoder in today's encoder-decoder architectures. Here, we look at differences between a recurrent-based encoder-decoder model with and without attention. We analyze the two model variants when tasked with a navigation instruction dataset designed to assess the compositional abilities of sequence-to-sequence models \citep{lake2018generalization}.

The key contributions of this work can be summarized as follows:
\begin{itemize}
  \item We introduce three new metrics for incrementality that help to understand the way that recurrent-based encoder-decoder models encode information;
  \item We conduct an in-depth analysis of how incrementally recurrent-based encoder-decoder models with and without attention encode sequential information;
  \item We confirm existing intuitions about attention-based recurrent models but also highlight some new aspects that explain their superiority over most attention-less recurrent models.
\end{itemize}

\section{Related Work}

Sequence-to-Sequence models that rely partly or fully on attention have gained much popularity in recent years (\citet{bahdanau2014neural}, \citet{vaswani2017attention}). Although this concept can be related to the prioritisation of information in the human visual cortex \citep{hassabis2017neuroscience}, it seems contrary to the incremental processing of information in a language context, as for instance recently shown empirically for the understanding of conjunctive generic sentences \citep{tessler2019incremental}.

In machine learning, the idea of incrementality has already played a role in several problem statements, such as inferring the tree structure of a sentence \citep{jacob2018learning}, parsing \citep{kohn2014incremental}, or in other problems that are naturally equipped with time constraints like real-time neural machine translation \citep{gu2016learning,dalvi2018incremental}, and speech recognition \citep{baumann2009assessing,jaitly2016online, graves2012sequence}.
Other approaches try to encourage incremental behavior implictly by modifying the model architecture or the training objective: \citet{guan2018story} introduce an encoder with an incremental self-attention scheme for story generation. \citet{wang2019deep} try to encourage a more incremental attention behaviour through masking for text-to-speech, while \citet{hupkes2018learning} guide attention by penalizing deviation from a target pattern.

The significance of the encoding process in sequence-to-sequence models has also been studied extensively by \citet{conneau2018you}. Proposals exploring how to improve the resulting approaches include adding additional loss terms \citep{serdyuk2017twin} or a second decoder \citep{jiang2018closed, korrel2019transcoding}.

\section{Metrics}\label{sec:metrics}

In this section, we present three novel metrics called \emph{Diagnostic Classifier Accuracy} (Section~\ref{sec:dc_acc}), \emph{Integration Ratio} (Section~\ref{sec:int_ratio}) and \emph{Representational Similarity} (Section~\ref{sec:rep_sim}) to assess the ability of models to process information incrementally. These metrics are later evaluated themselves in Section~\ref{sec:metric_comparison} and differ from traditional ones used to assess the incrementality of models, e.g. as the ones summarized by \citet{kohn2014incremental}, as they focus on the role of the encoder
in sequence-to-sequence models. It further should be noted that the ``optimal'' score of these measures with respect to downstream applications cannot defined explicity; they rather serve as a mean to uncover insights about the ways that attention changes a model's behavior, which might aid the development of new architectures.

\subsection{Diagnostic Classifier Accuracy}\label{sec:dc_acc}

Several works have utilized linear classifiers to predict the existence of certain features in the hidden activations\footnote{In this work, the terms \emph{hidden representation} and \emph{hidden activations} are used synonymously.} of deep neural networks \citep{hupkes2018visualisation,dalvi2019one,conneau2018you}. Here we follow the nomenclature of \citet{hupkes2018visualisation} and call these models
\emph{Diagnostic Classifiers} (DCs).

\begin{figure}[h]
    \centering
    \includegraphics[width=1.0\columnwidth]{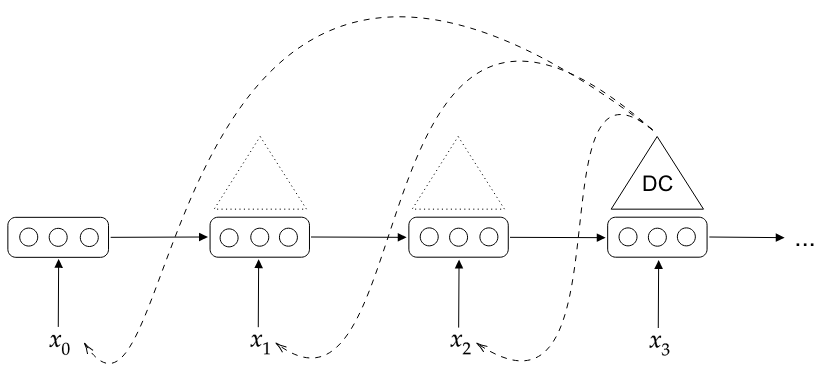}
    \caption{For the Diagnostic Classifier Accuracy, DCs are trained
    on the hidden activations to predict previously occuring tokens. The accuracies are averaged
    and potentially weighed by the distance between the hidden activations used for training the occurrence of the token to predict.}\label{fig:dcacc}
\end{figure}

We hypothesize that the hidden activations of an incremental model contain more information about previous
tokens inside the sequence. This is based on the assumption that attention-based models have no incentive to encode inputs recurrently, as previous representations
can always be revisited. To test this assumption, we train a DC on every time step $t > 1$ in a sequence
$t \in [1, \ldots T]$ to predict the $k$ most frequently occuring input tokens for all time steps $t^\prime < t$ (see Figure~\ref{fig:dcacc}).
For a sentence of length $T$, this results in $\sum_{t=2}^T \sum_{t^\prime=t-1}^t k$ trained DCs. To then generate
the corresponding training set for one of these classifiers, all activations from the network on a test set
are extracted and the corresponding tokens recorded. Next, all activations from time step $t$ are used as the training samples and all tokens to generate binary labels based on whether the target token $x_k$ occured on target time step $t^\prime$. As these data sets are highly unbalanced, class weights are also computed and used during training.

Applying this metric to a model, the accuracies of all classifiers after training are averaged on a given test set, which we call \emph{Diagnostic Classifier Accuracy} (DC Accuracy). We can test this way how much information about specific inputs is lost and whether that even matters for successful model performance, should it employ an encoding procedure of increasing abstraction like in \emph{Chunk-and-pass processing}. On the other hand, one might assume that a more powerful model might require to retain information about an input even if the same occured several time steps ago. To account for this fact, we introduce a modified version of this metric called \emph{Weighed Diagnostic Classifier Accuracy} (Weighed DC Accuracy), where we weigh the accuracy of a classifier based on the distance $t - t^\prime$.

\subsection{Integration Ratio}\label{sec:int_ratio}

\begin{figure}[]
    \centering
    \includegraphics[width=\columnwidth]{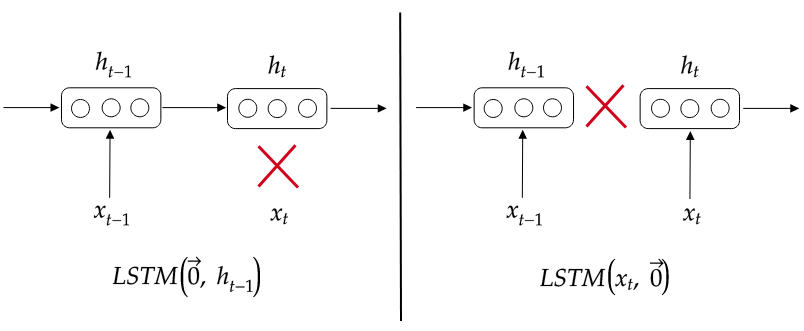}
    \caption{Illustration of a thought experiment about two types of extreme recurrent models. (\emph{Left}) The model completely ignores the current token and bases its new hidden state entirely on the previous one. (\emph{Right}) The model forgets the whole history and just encodes the current input.}\label{fig:intratio}
\end{figure}

Imagine an extreme attention-based model that does not encode information recurrently but whose hidden state $\bh_t$ is solely based on the current token $\bx_t$ (see right half of Figure~\ref{fig:intratio}). If we formalize an LSTM as a recurrent function $f_\theta: \mathbb{R}^n, \mathbb{R}^m \mapsto \mathbb{R}^m$ parameterized by weights $\theta$ that maps two continuous vector representations, in our case the $n$-dimensional representation of the current token $\bx_t \in \mathbb{R}^n$ and the $m$-dimensional previous hidden state representation $\bh_{t-1} \in \mathbb{R}^m$ to a new hidden state $\bh_t \in \mathbb{R}^m$, we can formalize the mentioned scenario as a recurrent function that completely ignores the pevious hidden state, which we can denote using a zero-vector $\vec{0} \in \mathbb{R}^m$: $\bh_t = f_\theta(\bx_t, \vec{0})$.

In a more realistic setting, we can exploit this thought experiment to quantify the amount of new information that is integrated into the current hidden representation by subtracting this hypothetical value from the actual value at timestep $t$:

\begin{equation}
    \Delta \bx_t = || \bh_t - f_\theta(\bx_t, \vec{0})||^2,
\end{equation}
where $||\ldots||^2$ denotes the $l_2$-norm. Conversely, we can quantify the amount of information that was lost from previous hidden states with:

\begin{equation}
    \Delta \bh_t = || \bh_t - f_\theta(\vec{0}, \bh_{t-1})||^2.
\end{equation}
In the case of the extreme attention-based model, we would expect $\Delta \bx_t = 0$, as no information from $\bh_{t-1}$ has been used in the transformation of $\bx_t$ by $f_\theta$. Likewise, the ``ignorant'' model would produce a value of $\Delta \bh_t = 0$, as any new hidden representation completely originates from a transformation of the previous one.

Using these two quanitities, we can formulate a metric expressing the average ratio between them throughout a sequence which we call \emph{Integration Ratio}:

\begin{equation}
  \phi_{int} = \frac{1}{T-1}{\displaystyle \sum_{t=2}^T}\frac{\Delta \bx_t}{\Delta \bh_t}
\end{equation}
This metric provides an intuitive insight into the (average) model behavior during the encoding process: For $\phi_{int} < 1$ it holds that $\Delta \bx_t < \Delta \bh_t$, signifying that the model prefers to integrate new information into the hidden state. Vice versa, $\phi_{int} > 1$ and therefore $\Delta \bx_t > \Delta \bh_t$ implies a preference to maintain a representation of preceding inputs, possibly at the cost of encoding the current token $\bx_t$ in an incomplete manner.

To account for the fact that integrating new information is more important at the beginning of a sequence -- as no inputs have been processed yet --
and maintaining a representation of the sentence is more plausile towards the end of a sentence, we introduce two linear
weighing terms with $\alpha_{\Delta \bx_t} = \frac{T - t}{T}$ and $\alpha_{\Delta \bh_t} = \frac{t}{T}$ for $\Delta \bx_t$ and $\Delta \bh_t$, respectively, which simplify to a single term $\alpha_t$:

\begin{equation}
  \phi_{int} = \frac{1}{Z}{\displaystyle \sum_{t=2}^T}\alpha_t\frac{\Delta \bx_t}{\Delta \bh_t} = \frac{1}{Z}{\displaystyle \sum_{t=2}^T}\frac{T-t}{t}\frac{\Delta \bx_t}{\Delta \bh_t},
\end{equation}
where $Z$ corresponds to a new normalizing factor such that $Z = \sum_{t=2}^T \frac{T-t}{t}$. It should be noted that the ideal score for this metric is unknown.
The motivation for this score merely lies in gaining inside into a model's behaviour, showing us whether it engages in a similar kind of \emph{eager processing} while having to handle memory constraints (in this case realized in the constant dimensionality of hidden representations) like in human cognition.

\subsection{Representational Similarity}\label{sec:rep_sim}

The sentences ``I saw a cat'' and ``I saw a feline'' only differ in terms of word choice, but essentially encode the same information. An incremental model, based on the Chunk-and-Pass processing described by \citet{christiansen2016now}, should arrive at the same or at least a similar, abstract encoding of these phrases.\footnote{In fact, given that humans built up sentence representations in a compositional manner, the same should hold for sentence pairs like ``I saw a cat'' and ''A feline was observed by me'', which is beyond the limits of the metric proposed here.} While the exact wording might be lost in the process, the information encoded should still describe an encounter with a feline creature. We therefore hypothesize that an incremental model should map the hidden activations of similar sequences of tokens into similar regions of the hidden activation space.
To test this assumption, we compare the representations produced by a model after encoding the same sequence of tokens - or \emph{history} - using their average pairwise
distance based on a distance measure like the $l_2$ norm or cosine similarity. We call the length of the history the \emph{order} of the \emph{Representational Similarity}.

\begin{figure}[h]
    \centering
    \includegraphics[width=\columnwidth]{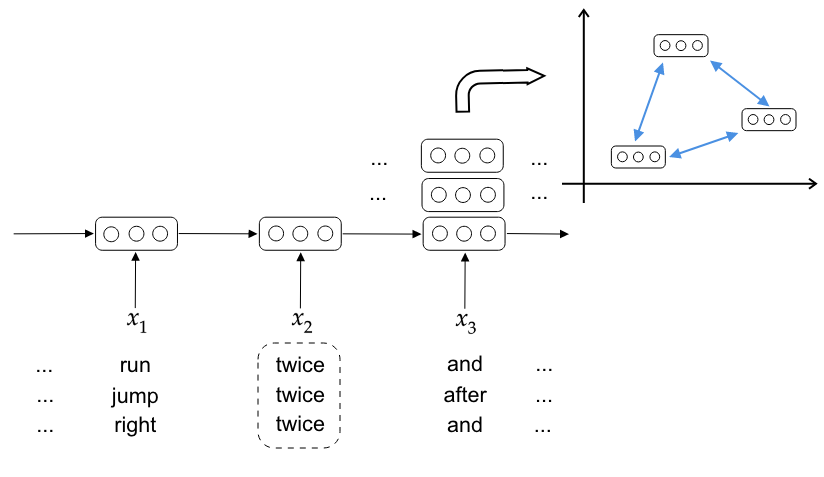}
    \caption{\emph{Representational Similarity} measures the average pair-wise distance of hidden representations after encoding the same subsquence of tokens (in this case the history is only of first order, i.e. $x_2$) as well as one arbitrary token $x_3$.}\label{fig:repsim}
\end{figure}

To avoid models to score high on this model metric by substituting most or all of a hidden representation with an encoding of the current token,\footnote{$\Delta \bx_t = 0$ in the framework introduced in the previous Section~\ref{sec:int_ratio}.} we only gather the hidden states for comparison after encoding another, arbitrary token (see Figure~\ref{fig:repsim}). We can therefore interpret the score as the ability to ``remember'' the same sequence of tokens in the past through the encoding.

The procedure is repeated for the $n$ most common histories of a specified order occuring in the test corpus over all time steps and, to obtain the final score, results are averaged.

\section{Setup}\label{sec:models}

We test our metric on two different architectures, trained on the SCAN dataset proposed by \citet{lake2018generalization}.
We explain both below.

\subsection{Data}\label{sec:data}

We use the SCAN data set proposed by \citet{lake2018generalization}: It is a simplified version of the
CommAI Navigation task, where the objective is to translate an order in natural language into a sequence
of machine-readable commands, e.g. ``jump thrice and look'' into \verb|I_JUMP I_JUMP I_JUMP I_LOOK|.
We focus on the \verb|add_prim_jump_split| \citep{loula2018rearranging}, where the model has to learn to generalize from seeing a command like
\verb|jump| only in primitive forms (i.e. by itself) to seeing it in composite forms during test time (e.g. \verb|jump twice|), where the remainder of the composite forms has been encountered in the context of other primitive commands during training.

The SCAN dataset has been proposed to assess the \emph{compositional} abilities of a model, which we believe to be deeply related with the concept of \emph{incrementality}, which is the target of our research.

\subsection{Models}

We test two seasoned architectures used in sequence processing, namely
a Long-Short Term Memory (LSTM) network \citep{hochreiter1997long} and an LSTM network with attention \citep{bahdanau2014neural}.
The attention mechanism creates a time-dependent context vector $\mathbf{c}_i$ for every decoder time step $i$ that is used together with the previous decoder hidden state. This vector is a weighted average of the output of the encoder, where the weights are calculated based on some sort of similarity measure. More specifically, we first calculate the energy $e_{it}$ between the last decoder hidden state $\mathbf{s}_{i-1}$ and any encoder hidden state $\bh_t$ using some function $a(\cdot)$

\begin{equation}
	e_{it} = a(\mathbf{s}_{i-1}, \mathbf{h}_t)
\end{equation}
We then normalize the energies using the softmax function and use the normalised attention weights $\alpha_{it}$ to create the context vector $\mathbf{c}_t$:



\begin{equation}
	\mathbf{c}_i = {\displaystyle \sum_{t=1}^T} \alpha_{it}\mathbf{h}_t
\end{equation}
In this work, we use a simple attention function, namely a dot product $a_{dot}$:

\begin{equation}\label{eq:attn-energy}
	a_{dot}(\mathbf{s}_{i-1}, \mathbf{h}_t) = \mathbf{s}_{i-1}^T\mathbf{h}_t,
\end{equation}
matching the setup originally introduced by \citet{bahdanau2014neural}.

\subsection{Training}
For both architectures, we train 15 single-layer uni-directional models, with an embedding and hidden layer size of 128.
We use the same hyperparameters for both architectures, to ensure compatibility.
More specifically, both models were trained for $50$ epochs using the Adam optimizer \citep{kingma2014adam} with the AMSgrad correction \citep{reddi2018convergence} and a learning rate of $0.001$ and a batch size of $128$.


\section{Results}

\begin{figure}
  \centering
  \includegraphics[width=0.85\columnwidth]{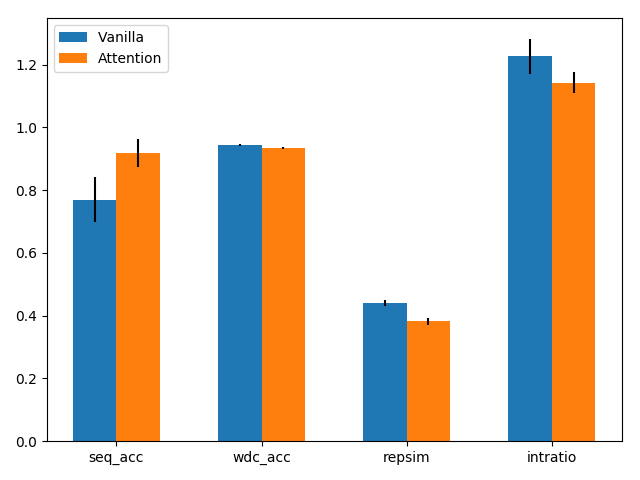}

  \caption{Results on SCAN \texttt{add\_prim\_left} with $n=15$. Abbreviations stand for sequence accuracy, weighed diagnostic classifier accuracy, integration ratio and representational similarity, respectively. All differences are statistically significant (using a Student's t-test with $p=0.05$).}\label{fig:results}
\end{figure}

We compute metric values for all 30 models (15 per architecture) that resulted from the training procedure described above.\footnote{The code used in this work is available online under \url{https://github.com/i-machine-think/incremental_encoding}.}
We plot the metric values, averaged over all runs for both models, in Figure~\ref{fig:results}.
For the representational similarity score, we use all instances of the $n=5$ most frequently occuring histories of length $2$ at all available time steps.
The unweighted DC accuracies are not depicted, as they do not differ substantially from their weighted counter part, for which we also try to detect the $k=5$ most frequently occuring inputs at every time step.


\subsection{Metric scores}\label{sec:results}


As expected, the standard attention model significantly outperforms the vanilla model in terms of sequence accuracy.
Surprisingly, both models perform very similarly in terms of weighed DC accuracy.
While one possible conclusion is that both models display a similar ability to store information about past tokens, we instead hypothesize that this can be explained by the fact that all sequences in our test set are fairly short ($6.8$ tokens on average).
Therefore, it is easy for both models to store information about tokens over the entire length of the input even under the constrained capacity of the hidden representations.
Bigger differences might be observed on corpora that contain longer sequences.


From the integration ratio scores (last column in Figure~\ref{fig:results}), it seems that, while both models prefer to maintain a history of previous tokens, the attention-based model contains a certain bias to add new information about the current input token.
This supports our suspicion that this model is less incentivized to build up expressive representations over entire sequences, as the encoder representation can always be revisited later via the attention mechanism.
Counterintuitively and perhaps surprisingly, it appears that the attention model produces representations that are more similar than the vanilla model, judging from the representational similarity score. To decode successfully, the vanilla model has to include information about the entire input sequence in the last encoder hidden state, making the encodings of similar subsequences more distinct because of their different prefixes.\footnote{Remember that to obtain these scores, identical subsequences of only length $2$ were considered.} In contrast, the representations of the attention model is able to only contain information about the most recent tokens, exclusively encoding the current input at a given time step in the extreme case, as the attention mechanism can select the required representations on demand. These results will be revisited in more detail in section \ref{sec:qualitative}.

\subsection{Metrics Comparison}\label{sec:metric_comparison}

\begin{figure}
  \centering
  \includegraphics[width=1.0\columnwidth]{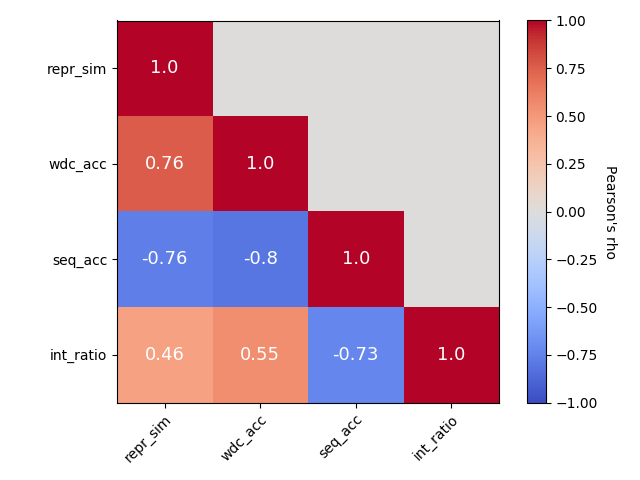}
  \caption{Correlations between metrics as heatmap of Pearson's rho values. $1$ indicates a strong positive correlation, $-1$ a negative one. Abbreviations correspond to the same metrics as in Figure~\ref{fig:results}. Best viewed in color.}
  \label{fig:metric-correlation}
\end{figure}

To further understand the salience of our new metrics, we use \emph{Pearson's correlation coefficient} to show their correlation with each
other and with sequence accuracy.
A heat map showing Pearson's $\rho$ values between all
metric pairs is given in Figure~\ref{fig:metric-correlation}.

We can observe that representational similarity and weighed DC accuracy display a substantial negative correlation with sequence accuracy. In the first case, this implies that the more similar representations of the same subsequences produced by the model's encoder are, the better the model itself performs later.\footnote{The representational similarity score actually expresses a degree of dissimilarity, i.e. a lower score results from more similar representations, therefore we identify a negative correlation here.} Surprisingly, we can infer from the latter case that storing more information about the previous inputs does not lead to better performance. At this point we should disentangle correlation from causation, as it is to be assumed that our hypothesis about the attention mechanism applies here as well: The attention is always able to revisit the encodings later during the decoding process, thus a hidden representation does not need to contain information about all previous tokens and the weighed DC accuracy suffers. Therefore, as the attention model performs better in terms of sequence accuracy, a negative correlation score is observed.
The same trend can be observed for the sequence accuracy - integration ratio pair, where the better performance of the attention model creates a significant negative correlation.

The last noteworthy observation can be found looking at the high positive correlation between the weighed DC accuracy and
representational similarity, which follows from the line of thought in Section~\ref{sec:results}: As the vanilla model has to squeeze information about the whole history into the hidden representation at every time step, encodings for a shorter subsequence become more distinct, while the attention model only encodes the few most recent inputs and are therefore able to produce more homogenous representations.

\subsection{Qualitative Analysis}\label{sec:qualitative}

\begin{figure}
    \centering
  \begin{subfigure}{\linewidth}
  \centering
  \includegraphics[width=1.0\linewidth]{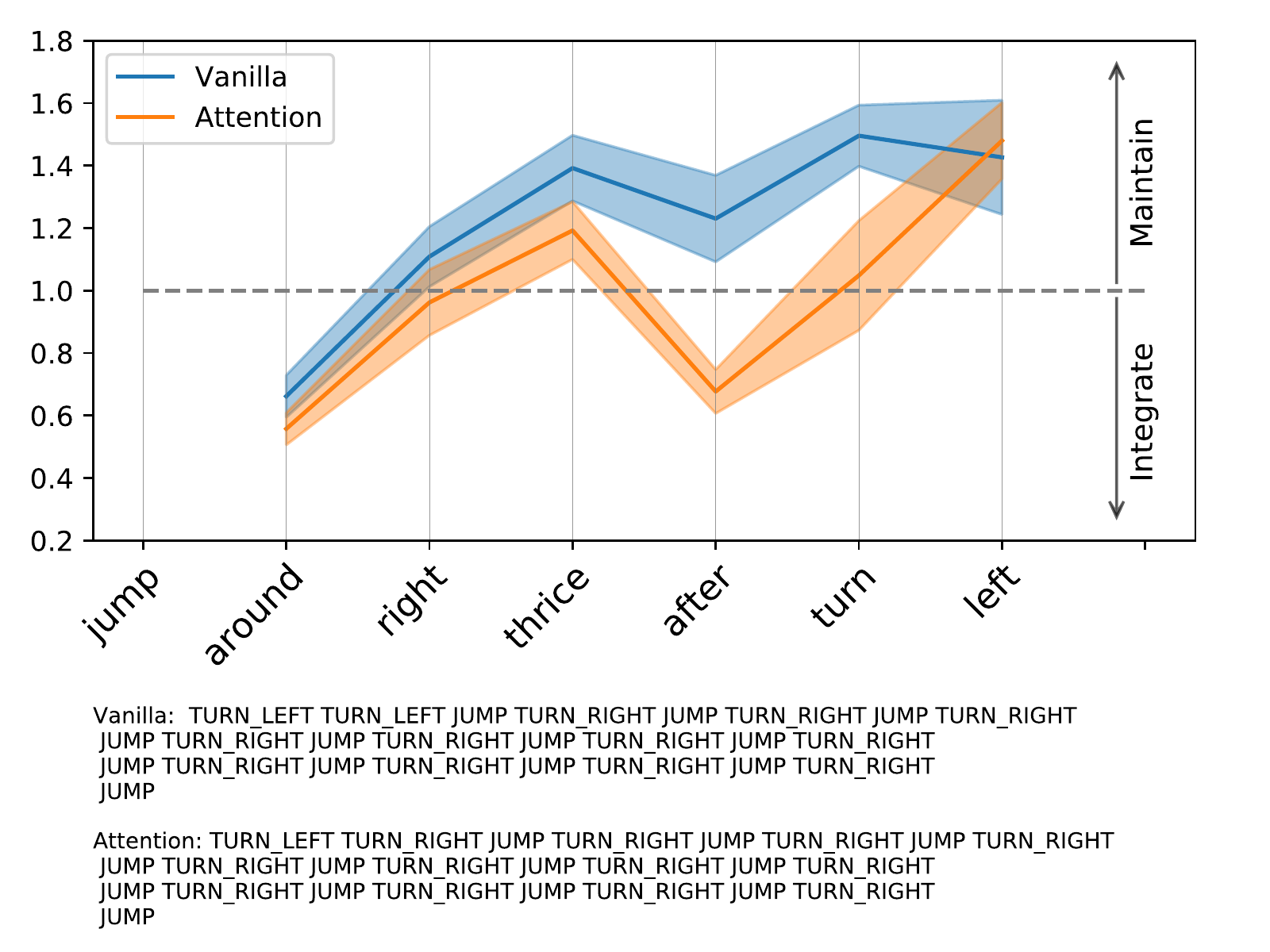}
  \caption{The vanilla model adds a redundant \texttt{TURN-LEFT} in the beginning.}\label{fig:qualitative_s1}
  \end{subfigure}

  \begin{subfigure}{\linewidth}
  \centering
  \includegraphics[width=\linewidth]{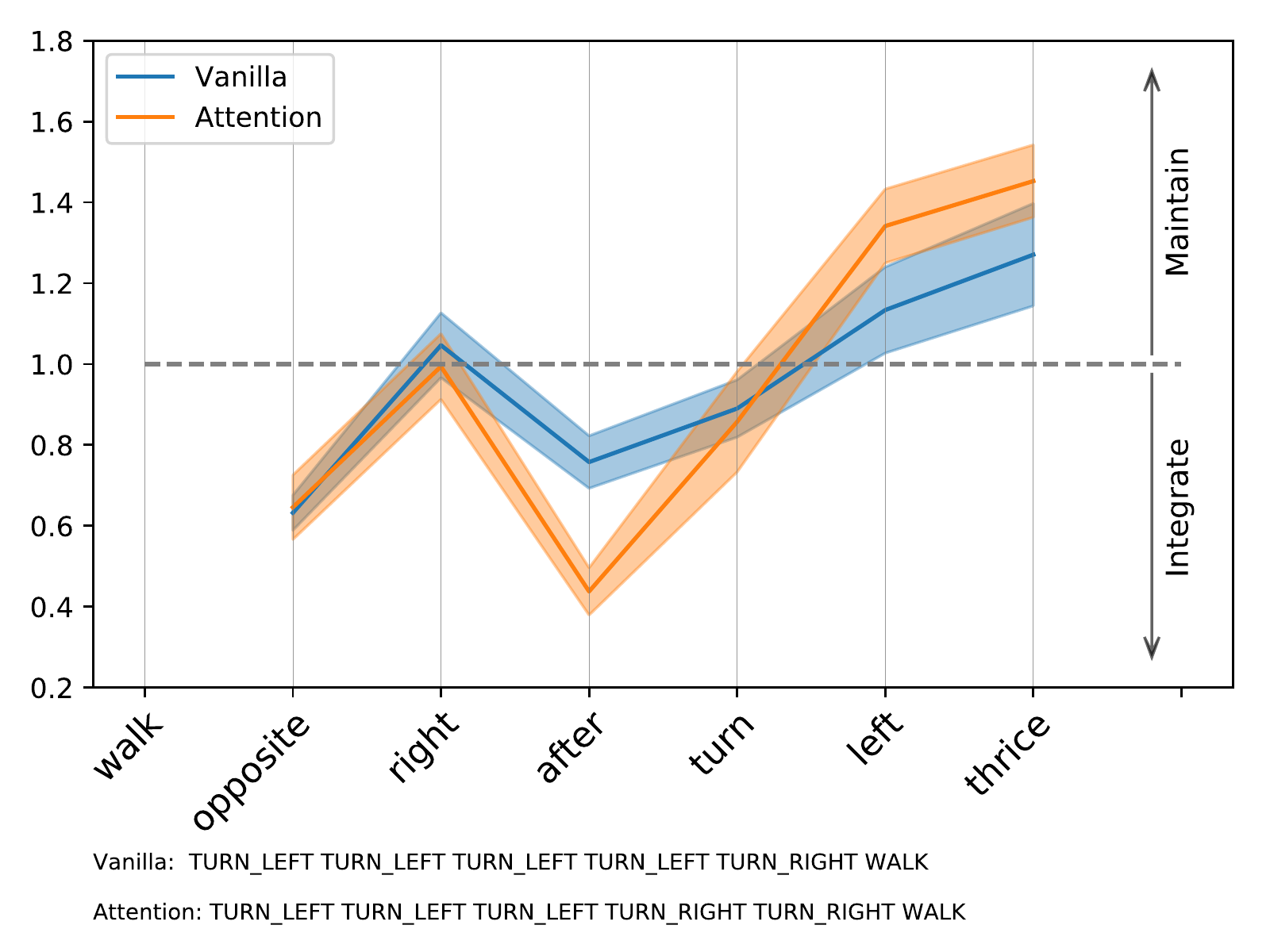}

  \caption{The vanilla model confuses left and right when decoding opposite.}\label{fig:qualitative_s2}
  \end{subfigure}

  \begin{subfigure}{\linewidth}
  \centering
  \includegraphics[width=\linewidth]{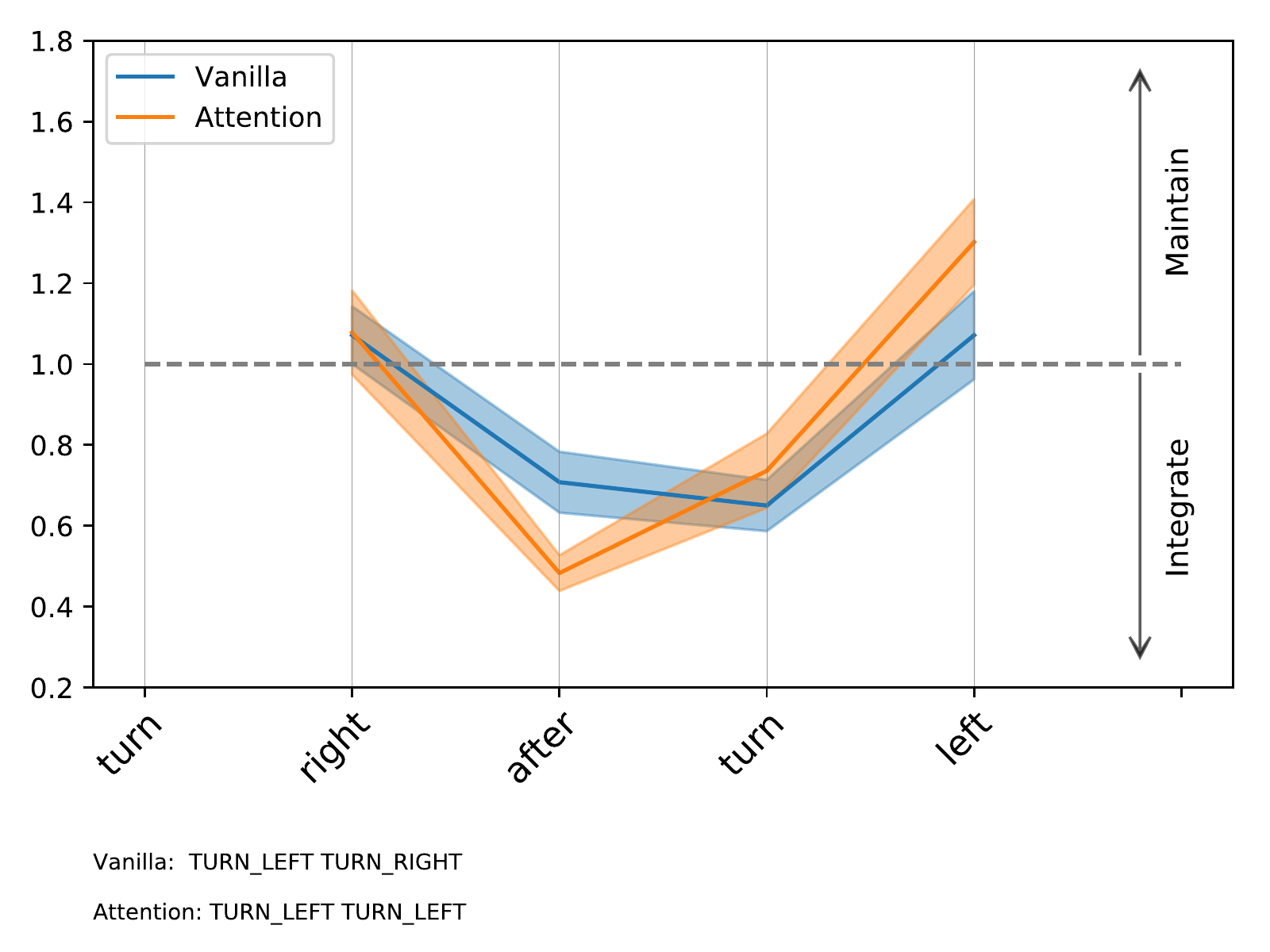}
  \caption{The attention model fails on a trivial sequence.}\label{fig:qualitative_s3}
  \end{subfigure}
  \caption{Qualitative analysis about the models' encoding behavior. Bounds show the standard deviation of integration ratio scores per time step. Decoded sentences are produced by having each model decode the sequence individually and then consolidating the solution via a majority vote. Resulting sequences have been slightly simplified for readability. Best viewed in color.}\label{fig:qualitative}
\end{figure}

We scrutinize the models' behavior when processing the same sequence by recording the integration ratio per time step and
contrasting them in plots, which are shown in Figure~\ref{fig:qualitative}. Figure~\ref{fig:qualitative_s1}~and~\ref{fig:qualitative_s2} are thereby indicative of a trend which further reinforces our hypothesis
about the behavior of attention-based models: As the orange curve lies below the vanilla model's blue curve in the majority of cases, we can even infer on a case by case basis
that these models tend to integrate more information at every time step than a vanilla LSTM. Interestingly, these distinct behaviors when processing information do not always lead to the models finding different solutions. 
In Figure~\ref{fig:qualitative} however, we present three error cases in which the models' results do diverge.

In Figure~\ref{fig:qualitative_s1}, we can see that the vanilla model decodes a second and redundant \verb|TURN-LEFT| in the beginning of the sequence. Although this happens right at the start, the corresponding part in the input sequence is actually encountered right at the end of the encoding process in the form of ``turn left'', where ``after'' in front of it constitutes an inversion of the sequence of operations. Therefore, when the vanilla model
starts decoding based on the last encoder hidden state, ``left'' is actually the most recently encoded token. We might assume that, due to this reason, the vanilla model might contain some sort of recency bias, which seems to corrupt some count information and leads to a duplicate in the output sequence. The attention model seems to be able to avoid this issue by erasing a lot of its prior encoded information when processing ``after'', as signified by the drop in the graph. Afterwards, only very little information seems to be integrated by the model.

The vanilla model commits a slightly different error in Figure~\ref{fig:qualitative_s2}: After both models decode three \verb|TURN-LEFT| correctly, it choses to decode ``opposite'' as \verb|TURN-LEFT| \verb|TURN-RIGHT| in contrast to the corect \verb|TURN-RIGHT| \verb|TURN-RIGHT| supplied by the attention model. It is to be assumed here that the last half of the input, ``turn left thrice'' had the vanilla model overwrite some critical information about the initial command. Again, the attention model is able to evade this problem by erasing a lot of its representation when encoding ``after'' and can achieve a correct decoding this critical part by attending to the representation produced at ``right'' later. ``turn left thrice'' can followingly be encoded without having to loose any past information.

Lastly, we want to shed some light on one of the rare failure cases of the \emph{attention model}, as given in Figure~\ref{fig:qualitative_s3}. Both models display very similar behavior when encoding this trivial sequence, yet only the vanilla model is able to decode it correctly. A possible reason for this could be found in the model's energy function: When deciding which encoded input to attend to for the next decoding step, the model scores potential candidates based on the last decoder hidden state (see eq. \ref{eq:attn-energy}), which was decoded as \verb|TURN-LEFT|. Therefore the most similar inputs token might appear to be \verb|TURN-LEFT| as well. Notwithstanding this explanation, it falls short of giving a conclusive reason why the model does not err in similar ways in other examples.

Looking at all three examples, it should furthermore be noted that the encoder of the attention model seems to anticipate the mechanism's behavior and learns to erase much of its representation after encoding one contiguous chunk of information, as exemplified by the low integration ratio after finishing the first block of commands in an input sequence. This freedom seems to enable the encoder to come up with more homogenous representations, i.e. that no information has to be overwritten and possibly being corrupted to process later, less related inputs, which also explains the lower representational similarity score in \ref{sec:results}.

\section{Conclusion}\label{sec:discussion}

In this work, we introduced three novel metrics that try to shine a light on the incremental abilities of the encoder in a sequence-to-sequence model and tested them on a LSTM-RNN with and without an attention mechanism. We showed how these metrics relate to each other and how they can be employed to better understand the encoding behavior of models and how these difference lead to performance improvements in the case of the attention-based model.

We confirm the general intuition that using an attention mechanism, due to its ability to operate on the whole encoded input sequence, prefers to integrate new information about the current token and is less pressured to maintain a representation for the whole input sequence, which seems to lead to some corruptions of the encoded information in case of the vanilla model. Moreover, our qualitative analysis suggests that the encoder of the attention model learns to chunk parts of the input sequence into
salient blocks, a behavior that is reminiscent of the Chunk-and-Pass processing described by \citet{christiansen2016now} and one component that is hypothesized to enable incremental processing in humans. In this way, the attention model most surprisingly seems to display a more incremental way of processing than the vanilla model.

These results open up several lines of future research: Although we tried to assess incrementality in sequence-to-sequence models in a quantitative manner, the notion of incremental processing lacks a formal definition within this framework. Thus, such definition could help to confirm our findings and aid in developing more incremental
architectures. It furthermore appears consequential to extend this methodology to deeper models and other RNN-variants as well as other data sets in order to confirm this work's findings.

Although we were possibly able to identify one of the components that build the foundation of human language processing \citep[as defined by][]{christiansen2016now} in attention models, more work needs to be done to understand how these dynamics play out in models that solely rely on attention like the Transformer \citep{vaswani2017attention} and how the remaining components could be realized in future models.

Based on these reflections, future work should attack this problem from a solid foundation: A formalization of incrementality in the context of sequence-to-sequence modelling could help to develop more expressive metrics. These metrics in turn could then
be used to assess possible incremental models in a more unbiased way.
Further thought should also be given to a fairer comparison of candidate models to existing baselines: The attention mechanism by \citet{bahdanau2014neural} and models like the Transformer operate without the temporal and memory pressure that is claimed to fundamentally shape human cognition \citet{christiansen2016now}. Controlling for this factor, it can be better judged whether incremental processing has a positive impact on the model's performance.
We hope that these steps will lead to encoders that create richer representations that can followingly be used back in regular sequence-to-sequence modelling tasks.

\section*{Acknowledgements}
DH is funded by the Netherlands Organization for Scientific Research (NWO),
through a Gravitation Grant 024.001.006 to the Language in Interaction Consortium. EB is funded by the European Union's Horizon 2020 research and innovation program under the Marie Sklodowska-Curie grant agreement No 790369 (MAGIC).

\bibliography{refs}

\begin{thebibliography}{26}
\expandafter\ifx\csname natexlab\endcsname\relax\def\natexlab#1{#1}\fi

\bibitem[{Bahdanau et~al.(2015)Bahdanau, Cho, and Bengio}]{bahdanau2014neural}
Dzmitry Bahdanau, Kyunghyun Cho, and Yoshua Bengio. 2015.
\newblock \href {http://arxiv.org/abs/1409.0473} {Neural machine translation by
  jointly learning to align and translate}.
\newblock In \emph{3rd International Conference on Learning Representations,
  {ICLR} 2015, San Diego, CA, USA, May 7-9, 2015, Conference Track
  Proceedings}.

\bibitem[{Baumann et~al.(2009)Baumann, Atterer, and
  Schlangen}]{baumann2009assessing}
Timo Baumann, Michaela Atterer, and David Schlangen. 2009.
\newblock Assessing and improving the performance of speech recognition for
  incremental systems.
\newblock In \emph{Proceedings of Human Language Technologies: The 2009 Annual
  Conference of the North American Chapter of the Association for Computational
  Linguistics}, pages 380--388. Association for Computational Linguistics.

\bibitem[{Christiansen and Chater(2016)}]{christiansen2016now}
Morten~H Christiansen and Nick Chater. 2016.
\newblock The now-or-never bottleneck: A fundamental constraint on language.
\newblock \emph{Behavioral and Brain Sciences}, 39.

\bibitem[{Conneau et~al.(2018)Conneau, Kruszewski, Lample, Barrault, and
  Baroni}]{conneau2018you}
Alexis Conneau, Germ{\'{a}}n Kruszewski, Guillaume Lample, Lo{\"{\i}}c
  Barrault, and Marco Baroni. 2018.
\newblock \href {https://aclanthology.info/papers/P18-1198/p18-1198} {What you
  can cram into a single {\textbackslash}{\textdollar}{\&}!{\#}* vector:
  Probing sentence embeddings for linguistic properties}.
\newblock In \emph{Proceedings of the 56th Annual Meeting of the Association
  for Computational Linguistics, {ACL} 2018, Melbourne, Australia, July 15-20,
  2018, Volume 1: Long Papers}, pages 2126--2136.

\bibitem[{Dalvi et~al.(2018{\natexlab{a}})Dalvi, Durrani, Sajjad, and
  Vogel}]{dalvi2018incremental}
Fahim Dalvi, Nadir Durrani, Hassan Sajjad, and Stephan Vogel.
  2018{\natexlab{a}}.
\newblock \href {https://aclanthology.info/papers/N18-2079/n18-2079}
  {Incremental decoding and training methods for simultaneous translation in
  neural machine translation}.
\newblock In \emph{Proceedings of the 2018 Conference of the North American
  Chapter of the Association for Computational Linguistics: Human Language
  Technologies, NAACL-HLT, New Orleans, Louisiana, USA, June 1-6, 2018, Volume
  2 (Short Papers)}, pages 493--499.

\bibitem[{Dalvi et~al.(2018{\natexlab{b}})Dalvi, Durrani, Sajjad, and
  Vogel}]{dalvi2019one}
Fahim Dalvi, Nadir Durrani, Hassan Sajjad, and Stephan Vogel.
  2018{\natexlab{b}}.
\newblock \href {https://aclanthology.info/papers/N18-2079/n18-2079}
  {Incremental decoding and training methods for simultaneous translation in
  neural machine translation}.
\newblock In \emph{Proceedings of the 2018 Conference of the North American
  Chapter of the Association for Computational Linguistics: Human Language
  Technologies, NAACL-HLT, New Orleans, Louisiana, USA, June 1-6, 2018, Volume
  2 (Short Papers)}, pages 493--499.

\bibitem[{Graves(2012)}]{graves2012sequence}
Alex Graves. 2012.
\newblock \href {http://arxiv.org/abs/1211.3711} {Sequence transduction with
  recurrent neural networks}.
\newblock \emph{CoRR}, abs/1211.3711.

\bibitem[{Guan et~al.(2018)Guan, Wang, and Huang}]{guan2018story}
Jian Guan, Yansen Wang, and Minlie Huang. 2018.
\newblock \href {http://arxiv.org/abs/1808.10113} {Story ending generation with
  incremental encoding and commonsense knowledge}.
\newblock \emph{CoRR}, abs/1808.10113.

\bibitem[{Hassabis et~al.(2017)Hassabis, Kumaran, Summerfield, and
  Botvinick}]{hassabis2017neuroscience}
Demis Hassabis, Dharshan Kumaran, Christopher Summerfield, and Matthew
  Botvinick. 2017.
\newblock Neuroscience-inspired artificial intelligence.
\newblock \emph{Neuron}, 95(2):245--258.

\bibitem[{Hochreiter and Schmidhuber(1997)}]{hochreiter1997long}
Sepp Hochreiter and J{\"u}rgen Schmidhuber. 1997.
\newblock Long short-term memory.
\newblock \emph{Neural computation}, 9(8):1735--1780.

\bibitem[{Hupkes et~al.(2019)Hupkes, Singh, Korrel, Kruszewski, and
  Bruni}]{hupkes2018learning}
Dieuwke Hupkes, Anand Singh, Kris Korrel, Germ{\'{a}}n Kruszewski, and Elia
  Bruni. 2019.
\newblock Learning compositionally through attentive guidance.
\newblock In \emph{CICLing: International Conference on Computational
  Linguistics and Intelligent Text Processing}.

\bibitem[{Hupkes et~al.(2018)Hupkes, Veldhoen, and
  Zuidema}]{hupkes2018visualisation}
Dieuwke Hupkes, Sara Veldhoen, and Willem Zuidema. 2018.
\newblock Visualisation and'diagnostic classifiers' reveal how recurrent and
  recursive neural networks process hierarchical structure.
\newblock \emph{Journal of Artificial Intelligence Research}, 61:907--926.

\bibitem[{Jacob et~al.(2018)Jacob, Lin, Sordoni, and
  Bengio}]{jacob2018learning}
Athul~Paul Jacob, Zhouhan Lin, Alessandro Sordoni, and Yoshua Bengio. 2018.
\newblock Learning hierarchical structures on-the-fly with a
  recurrent-recursive model for sequences.
\newblock In \emph{Proceedings of The Third Workshop on Representation Learning
  for NLP}, pages 154--158.

\bibitem[{Jaitly et~al.(2016)Jaitly, Le, Vinyals, Sutskever, Sussillo, and
  Bengio}]{jaitly2016online}
Navdeep Jaitly, Quoc~V Le, Oriol Vinyals, Ilya Sutskever, David Sussillo, and
  Samy Bengio. 2016.
\newblock An online sequence-to-sequence model using partial conditioning.
\newblock In \emph{Advances in Neural Information Processing Systems}, pages
  5067--5075.

\bibitem[{Jiang and Bansal(2018)}]{jiang2018closed}
Yichen Jiang and Mohit Bansal. 2018.
\newblock \href {https://aclanthology.info/papers/D18-1440/d18-1440}
  {Closed-book training to improve summarization encoder memory}.
\newblock In \emph{Proceedings of the 2018 Conference on Empirical Methods in
  Natural Language Processing, Brussels, Belgium, October 31 - November 4,
  2018}, pages 4067--4077.

\bibitem[{Kingma and Ba(2015)}]{kingma2014adam}
Diederik~P. Kingma and Jimmy Ba. 2015.
\newblock \href {http://arxiv.org/abs/1412.6980} {Adam: {A} method for
  stochastic optimization}.
\newblock In \emph{3rd International Conference on Learning Representations,
  {ICLR} 2015, San Diego, CA, USA, May 7-9, 2015, Conference Track
  Proceedings}.

\bibitem[{K{\"o}hn and Menzel(2014)}]{kohn2014incremental}
Arne K{\"o}hn and Wolfgang Menzel. 2014.
\newblock Incremental predictive parsing with turboparser.
\newblock In \emph{Proceedings of the 52nd Annual Meeting of the Association
  for Computational Linguistics (Volume 2: Short Papers)}, volume~2, pages
  803--808.

\bibitem[{Korrel et~al.(2019)Korrel, Hupkes, Dankers, and
  Bruni}]{korrel2019transcoding}
Kris Korrel, Dieuwke Hupkes, Verna Dankers, and Elia Bruni. 2019.
\newblock Transcoding compositionally: using attention to find more
  generalizable solutions.
\newblock \emph{BlackboxNLP 2019, ACL}.

\bibitem[{Lake and Baroni(2018)}]{lake2018generalization}
Brenden Lake and Marco Baroni. 2018.
\newblock Generalization without systematicity: On the compositional skills of
  sequence-to-sequence recurrent networks.
\newblock In \emph{International Conference on Machine Learning}, pages
  2879--2888.

\bibitem[{Loula et~al.(2018)Loula, Baroni, and Lake}]{loula2018rearranging}
Jo{\~a}o Loula, Marco Baroni, and Brenden~M Lake. 2018.
\newblock Rearranging the familiar: Testing compositional generalization in
  recurrent networks.
\newblock \emph{arXiv preprint arXiv:1807.07545}.

\bibitem[{Neubig et~al.(2017)Neubig, Cho, Gu, and Li}]{gu2016learning}
Graham Neubig, Kyunghyun Cho, Jiatao Gu, and Victor O.~K. Li. 2017.
\newblock \href {https://aclanthology.info/papers/E17-1099/e17-1099} {Learning
  to translate in real-time with neural machine translation}.
\newblock In \emph{Proceedings of the 15th Conference of the European Chapter
  of the Association for Computational Linguistics, {EACL} 2017, Valencia,
  Spain, April 3-7, 2017, Volume 1: Long Papers}, pages 1053--1062.

\bibitem[{Reddi et~al.(2018)Reddi, Kale, and Kumar}]{reddi2018convergence}
Sashank~J Reddi, Satyen Kale, and Sanjiv Kumar. 2018.
\newblock On the convergence of adam and beyond.

\bibitem[{Serdyuk et~al.(2018)Serdyuk, Ke, Sordoni, Trischler, Pal, and
  Bengio}]{serdyuk2017twin}
Dmitriy Serdyuk, Nan~Rosemary Ke, Alessandro Sordoni, Adam Trischler, Chris
  Pal, and Yoshua Bengio. 2018.
\newblock \href {https://openreview.net/forum?id=BydLzGb0Z} {Twin networks:
  Matching the future for sequence generation}.
\newblock In \emph{6th International Conference on Learning Representations,
  {ICLR} 2018, Vancouver, BC, Canada, April 30 - May 3, 2018, Conference Track
  Proceedings}.

\bibitem[{Tessler et~al.(2019)Tessler, Gu, and Levy}]{tessler2019incremental}
Michael~Henry Tessler, Karen Gu, and Roger~Philip Levy. 2019.
\newblock Incremental understanding of conjunctive generic sentences.

\bibitem[{Vaswani et~al.(2017)Vaswani, Shazeer, Parmar, Uszkoreit, Jones,
  Gomez, Kaiser, and Polosukhin}]{vaswani2017attention}
Ashish Vaswani, Noam Shazeer, Niki Parmar, Jakob Uszkoreit, Llion Jones,
  Aidan~N Gomez, {\L}ukasz Kaiser, and Illia Polosukhin. 2017.
\newblock Attention is all you need.
\newblock In \emph{Advances in Neural Information Processing Systems}, pages
  5998--6008.

\bibitem[{Wang(2019)}]{wang2019deep}
Gary Wang. 2019.
\newblock \href {http://arxiv.org/abs/1903.07398} {Deep text-to-speech system
  with seq2seq model}.
\newblock \emph{CoRR}, abs/1903.07398.

\end{thebibliography}
\bibliographystyle{acl_natbib}

\end{document}